\documentclass[10pt,twocolumn,letterpaper]{article}

\usepackage{iccv}
\usepackage{times}
\usepackage{epsfig}
\usepackage{graphicx}
\usepackage{amsmath}
\usepackage{amssymb}
\usepackage{color}
\usepackage{textcomp}
\usepackage[font=small,labelsep=period]{caption}
\usepackage{subfloat}
\usepackage{subfig}
\usepackage[normalem]{ulem}

\usepackage[breaklinks=true,bookmarks=false]{hyperref}

\iccvfinalcopy

\definecolor{orange}{rgb}{1.0, 0.5, 0.0}

\ificcvfinal\fi
\begin{document}

\title{Learned Multi-Patch Similarity}

\author{Wilfried Hartmann\textsuperscript{1}\hspace{3em}Silvano
  Galliani\textsuperscript{1}\hspace{3em}Michal Havlena\textsuperscript{2}\\
  Luc Van Gool\textsuperscript{3,4}\hspace{3em}Konrad Schindler\textsuperscript{1}\\ \\
  \textsuperscript{1}Photogrammetry and Remote Sensing, ETH Zurich, Switzerland \\
  \textsuperscript{2}Vuforia, PTC, Vienna, Austria \\
  \textsuperscript{3}CVL, ETH Zurich, Switzerland\hspace{3em}\textsuperscript{4}PSI, KU Leuven, Belgium
}

\maketitle
\thispagestyle{empty}

\begin{abstract}
  Estimating a depth map from multiple views of a scene is a
  fundamental task in computer vision. As soon as more than two
  viewpoints are available, one faces the very basic question how to
  measure similarity across $>$2 image patches. Surprisingly, no
  direct solution exists, instead it is common to fall back to more or
  less robust averaging of two-view similarities. Encouraged by the
  success of machine learning, and in particular convolutional neural
  networks, we propose to learn a matching function which directly
  maps multiple image patches to a scalar similarity
  score. Experiments on several multi-view datasets demonstrate that
  this approach has advantages over methods based on pairwise patch
  similarity.
\end{abstract}

\section{Introduction}

3D~reconstruction from two or more images of the same scene is a
central problem in computer vision. Assuming that the camera poses are
already known, the problem reduces to (multi-view) stereo matching,
\ie, establishing dense point correspondences between the images, which
can then be converted to 3D~points by triangulating the corresponding
rays.
The core of stereo matching itself is a function to measure the
similarity between points in different images, respectively between
the points surrounding image patches. Once such a \emph{similarity
  measure} is available, it can be computed for a list of different
putative correspondences to find the one with the highest similarity.

For the classic two-view stereo case, the definition of a similarity
function is comparatively straight-forward: transform the image
intensities of the two patches such that more similar ones end up
closer to each other, according to some pre-defined distance metric.
Many methods have been proposed, including simple sum-of-squared
differences, (inverse) normalized cross-correlation to afford
invariance against linear brightness changes, and even more robust
measures like the Hamming distance between Census descriptors.
More recently it has also been proposed to learn the distance metric
discriminatively from matching and non-matching training examples.

\begin{figure}[t]
\begin{center}
  \includegraphics[width=0.975\columnwidth]{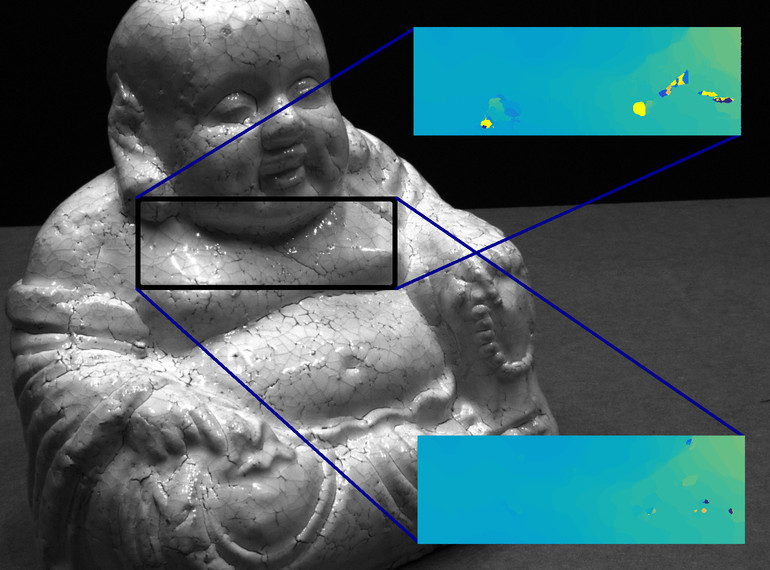}
\caption{Multi-view depth estimation. A conventional, pairwise similarity like ZNCC \emph{(top)} is unable to find the correct depth in corrupted regions, \eg~specular reflections; whereas the proposed multi-view similarity \emph{(bottom)} can predict correct depth values.}
\label{fig:teaser}
\end{center}
\end{figure}

In practice, multi-view stereo is often the method of choice, since
the higher redundancy and larger number of viewpoints allows for more
robust 3D~reconstruction.
A host of rather successful multi-view stereo methods exist (see
benchmark results such as~\cite{Jensen2014,Schoeps2017CVPR,seitz2006comparison, strecha2008}). Surprisingly, these methods in fact have no mechanism to
measure the similarity between $>$2 image patches that form a putative
multi-view correspondence.
Instead, they heuristically form a consensus over the pair-wise
similarities, or a subset of them (most often the similarities from
all other stereo partners to a ``reference image'' in which the depth
map is parametrized).
We note that the same is true for ``multi-view stereo'' methods that
reconstruct implicit~\cite{zach2008fast} or
explicit~\cite{delaunoy2008minimizing} surfaces. These either
reconstruct points from two images and delay the multi-view
integration to the surface fitting stage; or
they measure photo-consistency between pairs of images, or between
some ``mean'' surface texture and the individual images.

Here, we pose the question \emph{why not directly measure
multi-view similarity?} Encouraged by the successes of learned
similarity measures, we propose a multi-stream (``Siamese'')
convolutional neural network architecture that takes as input a number
of $n>2$ image patches, and outputs a scalar similarity score.
The network is learned directly from matching and non-matching example
patches, keeping decisions like the right weighting of
individual images or image pairs (\eg, to account for varying
contrast) or the robustness of the consensus mechanism (\eg, due to
occlusions, specularities, and other disturbances in individual
images) implicit.
An alternative view of our work is as a multi-view extension of
learning-based stereo
correspondence~\cite{MooYi2016,Zagoruyko2015,Zbontar2015} to more than
two views. We posit that the learning-based stereo should profit from
the multi-view setup, precisely because the additional images afford
the necessary redundancy to detect and resolve situations where
two-view stereo struggles.
To test our similarity measure, we embed it into a standard depth
estimation setup, namely multi-view
plane-sweeping~\cite{Collins1996}: for each pixel in an image we
compute similarity scores for a range of depths along the ray, and
pick the depth with the highest similarity.

There are different strategies to cast stereo matching into a machine
learning problem. One can either fix the metric (\eg, Euclidean
distance) and learn to map image patches to ``descriptor vectors''
that, according to that metric, have small distance for matching
patches and large distance for non-matching patches~\cite{MooYi2016}.
For our case, that approach does not resolve the problem of defining an
$n$-view distance.
Alternatively, one can map raw patches to descriptors according to
some conventional recipe, \eg SIFT, and train a
similarity/dissimilarity metric between
them~\cite{yu2006toward}.
However, given the spectacular improvements that learned features from
CNNs have brought to computer vision, we prefer to learn the mapping
end-to-end from raw pixel intensities to a similarity score.

Conceptually, it is straight-forward to design a CNN for multi-view
similarity.
A standard architecture to jointly process two images with similar
image statistics are ``Siamese'' networks: the two inputs are passed
through identical streams with tied weights and then combined for the
final decision layers by simple addition or concatenation.
We do the same for $n>2$ images and set up $n$ parallel streams with
tied weights, without introducing additional free parameters.
The network is trained on a portion of the public DTU multi-view
dataset~\cite{Jensen2014}, and evaluated on the remaining part of it,
as well as on an unrelated public dataset. We will also show that it
is possible to vary the number $n$ of input image patches at test time
without retraining the network.
The comparison to other conventional and learning-based approaches
demonstrates the benefit of evaluating direct multi-view similarity,
especially in the case when the reference view is corrupted, \eg, due
to specular reflection -- see Figure~\ref{fig:teaser}.

\section{Related Work}

\paragraph{Depth Estimation.}
An enormous body of literature exists on stereo matching.
Much of the early work addressed the minimal two-view
case~\cite{Barnard1982}. Given the relative pose of the two images,
stereo correspondence reduces to a 1D~search along corresponding
epipolar lines. It is standard practice to impose a smoothness prior, both along
and across epipolar lines, to better cope with ambiguities of the local
similarity measure.
For disparity map computation a commonly used measure is the sum of
absolute differences (SAD)~\cite{Kanade1996}. Here the similarity
measure is based on intensity differences of two image patches with
the same size. Zero-mean normalized cross correlation
(ZNCC)~\cite{Hannah1974} is another popular similarity measure, which
features invariance against linear brightness changes. A discussion of
the complete two-view stereo literature is beyond the scope of this
paper, for an overview and further reading please refer to benchmarks
like the Middlebury Stereo page~\cite{Scharstein2002taxonomy} or
KITTI~\cite{geiger2012we}.

While not as over-whelming as in the two-view case, there still is an
extensive literature about multi-view stereo.
Conceptually, there are two main approaches to generate depth maps from
multiple views. One exhaustively tests for all possible depths, often
by ``sweeping'' a fronto-parallel plane through the 3D~object space
along the depth axis~\cite{Collins1996} or in multiple
directions~\cite{Haene2014} to sample different depth values
efficiently. The other avoids exhaustive testing and instead relies on
efficient random sampling and propagation schemes like
PatchMatch~\cite{barnes2009patchmatch} to find a good depth estimate
at every position, \eg~\cite{Galliani2015}.
For both strategies, efficient GPU implementations exist.
Again, we refer the reader to benchmark datasets such as
~\cite{seitz2006comparison} and~\cite{Jensen2014} for an overview of
contemporary multi-view stereo algorithms.
We note that there is also a large body of work termed ``multi-view
reconstruction'' that in fact computes depth maps from two views and
focuses on integrating the corresponding 3D~point clouds into
implicit~\cite{galliani2016just,zach2008fast} or explicit~\cite{delaunoy2008minimizing}
surface representations.
Since these methods only start \emph{after} having lifted image points
to 3D~scene space, they are not directly relevant for the present
paper.

\paragraph{Patch Similarity Learning.}

With the rise of machine learning for computer vision problems, it has
also been proposed to learn the similarity measure for (two-view)
stereo.
Early work still relied on hand-tuned descriptors such as SIFT, the
learning served to ``distort'' the descriptor space so that nearby
false matches get pushed apart and the distance becomes more
discriminative~\cite{yu2006toward}.
The advent of deep learning suggested that the bottleneck might be the
descriptors themselves rather than the distance metric, so it was
proposed to learn similarity directly from raw images~\cite{Zbontar2015}.
Closely related work started from separate steps for descriptor
learning and metric learning, and unified them to effectively
obtain a direct similarity prediction from raw image
data~\cite{Han2015} as well.
An extensive study of similarity measures based on different CNN
architectures is presented in~\cite{Zagoruyko2015}. That work also
showed that CNN-based similarities outperform both classical
descriptor spaces like SIFT~\cite{Lowe-IJCV-2004} and other learned
descriptors such as~\cite{Simonyan2014}.
Another strategy is to learn patch descriptors, but freeze the
distance measure used to compare them.
The recently proposed LIFT descriptor~\cite{Simo-Serra2015,MooYi2016}
is learned with the help of a Siamese network, using a loss function
that ensures that descriptors of matching patches end up having low
Euclidean distance, whereas non-matching descriptors have not.
The learned output is a 128-dimensional descriptor vector which corresponds to the size of the SIFT descriptor vector~\cite{Lowe-IJCV-2004} so that LIFT can serve as a drop-in
replacement for SIFT or similar hand-coded descriptors in existing
matching pipelines.

Yet, the learned descriptors still share the limitation of most
two-view stereo methods, that similarity is measured only for image
pairs, as a distance in descriptor space.

\section{Measuring $n$-way Patch Similarity with a Neural Network}

We aim for a function that directly maps $n>2$ image
patches $p_i$ to a scalar similarity score $S(p_1,p_2,\hdots, p_n)$.
Our proposed solution is to learn that function from example data, in
the spirit of what is sometimes called ``metric learning''.\footnote {We refrain from using that name, since the learned
  similarity score is not guaranteed to be a metric in the
  mathematical sense.} As learning engine, we use a convolutional neural network.
We point out that the strategy to learn a multi-patch similarity is
generic, and not limited to stereo correspondence. In fact, a main
argument for learning it is that a learned score can be tuned to
different applications, just by choosing appropriate training data.
In our target application, the $n$~patches are the reprojections of a
candidate 3D~point into $n$~different views of the scene.

For our purposes, we make the assumption that every set of patches in
the training data is either ``similar'' or ``dissimilar'', and do not
assign different degrees of similarity.
\Ie, we cast the similarity as a binary classification
problem. Conveniently, this means that the similarity score is bounded
between $0$~(maximally dissimilar) and $1$~(maximally similar).
Such a hard, discriminative approach reflects the situation of stereo
matching (and correspondence estimation in general), where for any
given pixel in a reference image there is only one correct answer,
namely the set of patches that correspond to the ground truth depth.
We note that for other applications, for example image retrieval, the
definition may not be suitable and would have to be replaced with a
more gradual, continuous one (for which it is however not as
straight-forward to generate training labels).

\begin{figure}[t]
\begin{center}
\includegraphics[width=0.8\columnwidth]{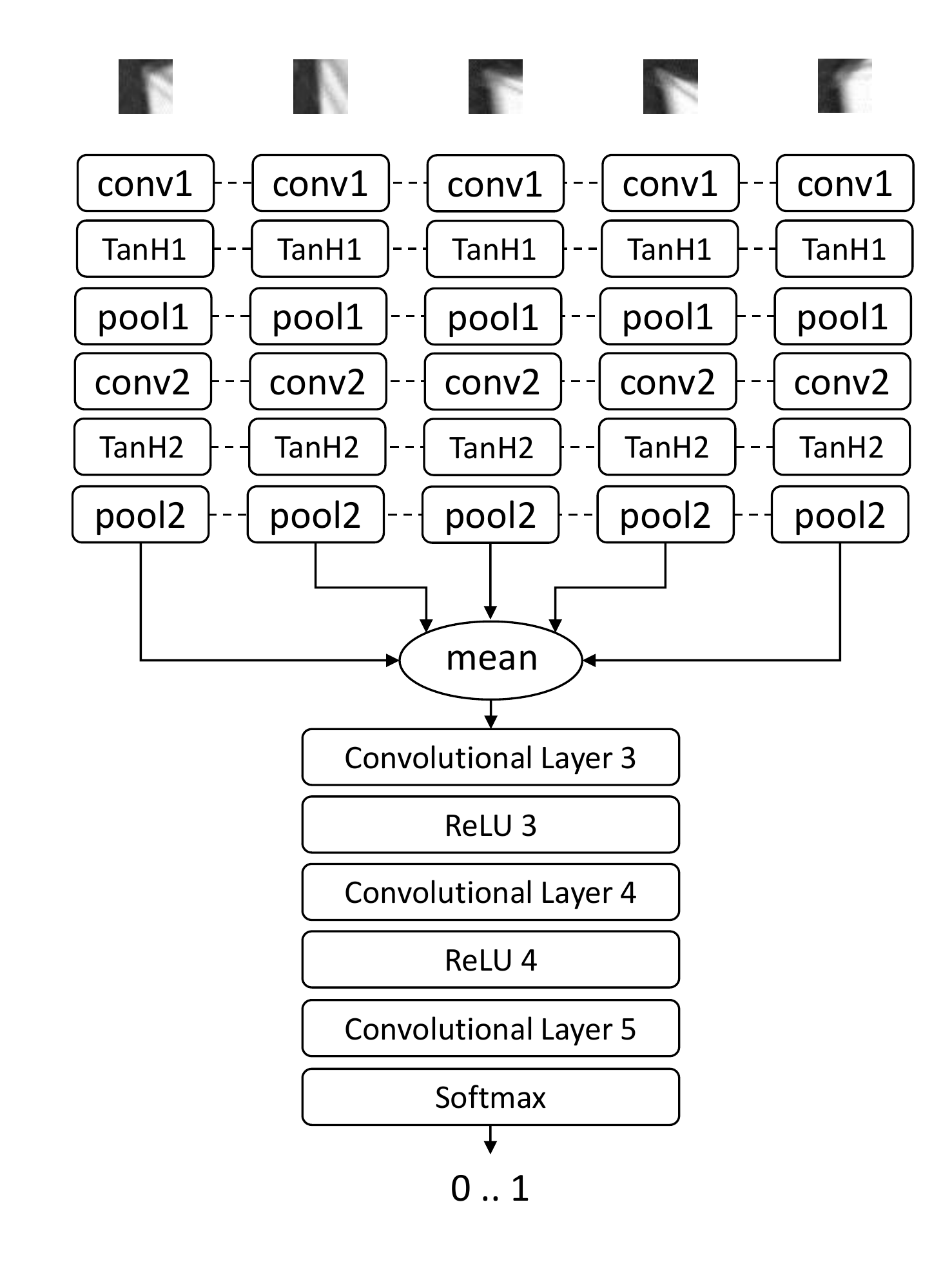}
\vspace{-1.5em}
\caption{The proposed network architecture with five ``Siamese'' branches. The input are the five patches, the output is the similarity score that is used to select the correct depth value }
\label{fig:cnn}
\end{center}
\vspace{-2em}
\end{figure}

\subsection{Network Architecture}

The network we use for learning a depth map is illustrated in
Figure~\ref{fig:cnn}. Its inputs are $n$~image patches (w.l.o.g.\ we set
$n=5$ for the remainder of the paper) of size $32\times32$ pixels.
The early layers process each patch separately with the same set of
weights, corresponding to an $n$-way Siamese network architecture.
Identical weights in the different branches are a natural choice,
since the result should not depend on the order of the input patches.
Note that the number of free weights to be learned is independent of
the number $n$ of views.

Each branch starts with a convolutional layer with $32$ filter kernels
of size $5\times5$. Then follow a $tanh$ non-linearity and a
$max$-pooling layer with kernel size $2\times2$.
That same sequence is then repeated, this time with $64$ filter
kernels of size $5\times5$, and $max$-pooling over $2\times2$ neurons.
From the resulting $5\times5\times64$ layers the mean is taken over all
$n$ branches. The output is passed through two convolutional layers
with $2048$ filter kernels, each followed by a $ReLU$ non-linearity,
and a final convolutional layer with $2$ filter kernels. The final
network output is the similarity score.

The described, fully convolutional architecture has the advantage that
it is much more efficient to compute for larger images at test time
than a per-pixel similarity.
Due to the two pooling layers, the network outputs a similarity score
for every $4\times4$ pixel region.

When designing the network, the straightforward approach would be to
concatenate the outputs of the individual branches, instead of
averaging them. We evaluate both options and find their performance on
par, see Tab.~\ref{tab:quantit}. Note that averaging the branch
outputs makes it possible to input a varying number of views without
retraining the network.

As usual, the exact network design is found empirically and is
somewhat hard to interpret.
We tested several different architectures and found the described one
to work best. Compared to other current architectures for image
analysis, our network needs fewer convolutional layers, presumably
because of the small input patch size; and, interestingly, old-school
$tanh$ non-linearities work better than $ReLU$ in the early layers.

\section{Application to Multi-view Stereo}

Having introduced the multi-view similarity computation, we use it as
a building block in a fairly standard multi-view stereo
pipeline~\cite{Haene2014}, assuming known camera poses (\eg, from
prior structure-from-motion computation).
As usual, one camera is designated as the reference view for which the depth
map is computed. We emphasize that, other than in most existing
multi-view approaches, the reference view serves \emph{only} to fix
the parametrization of the depth map.
In our method, its patches do not have a privileged role as the
``source'' templates to be combined to the ``target'' patches of the
stereo partners in a pairwise fashion.
Note, in a multi-view setup it can happen that points are occluded in
the reference view, but visible in several other views. In that case
the learned similarity score may assign the highest similarity to a
point that is invisible in the reference image.
If the final product of multi-view stereo is a 3D~point cloud, this
behavior does not hurt (except that the corresponding point on the
occluder is missing).

To find the most likely depth for a given pixel, we discretize the
depth along the viewing direction of the reference view, as in
standard plane-sweep stereo.
Matching then reduces to exhaustively testing all possible depth
values and choosing the one with the highest similarity.
For a given patch in the reference view the matching patches in the
other images are extracted. This is repeated for all planes, where each
plane corresponds to a discrete depth value. Each $n$-tuple of patches
(including the reference patch, which is the same for every depth) is
fed to the similarity network.
Not surprisingly, rather larger patches give more reliable similarity
estimates, but there is a trade-off  against memory consumption and
computational cost. We found $32\times32$ pixels
to be a good compromise.

\paragraph{Training the Network.}
\label{sec:train}

\begin{figure}[t]
\begin{center}
  \includegraphics[width=1.0\columnwidth]{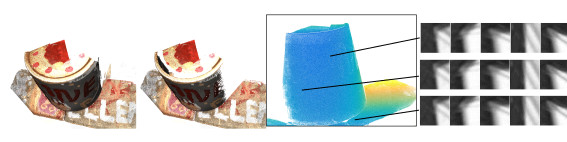}
\caption{Generating training data. The ground truth 3D~point cloud is
  processed using the visibility check of~\cite{Katz2007}. Points not
  visible in the reference view are removed. Next, points are randomly
  selected, projected into the image, and a patch centered at the
  projection is cropped out.}
\label{fig:patches}
\end{center}
\end{figure}

Our network is implemented in Caffe~\cite{Jia2014}, and learned from
scratch.
As training data, we sample image patches from 49~scenes of the DTU
dataset~\cite{Jensen2014} (Fig.~\ref{fig:patches}).

Positive samples (similarity $1$) are selected using the ground truth
depth information. Negative samples are selected by picking patches
from depth planes $15$ steps before or after the ground truth.
Note the power of the learning approach to optimally tune to the
application: patches on correct epipolar lines are the only ones the
network will ever see at test time (assuming correct camera poses).
Using only such patches for training ensures that the model is
``complete'', in the sense that it sees all sorts of patch
combinations that can occur along epipolar lines; but at the same time
it is also ``tight'' in the sense that it does not contain irrelevant
negatives that do not correspond to any valid set of epipolar
geometries and merely blur the discriminative power of the similarity.

It is a recurring question what proportion of positive and negative
samples to use in situations with very unbalanced class distributions
(for multi-view stereo, only one sample within the entire depth range
is positive for each pixel).
For our case it worked best to use a balanced training set in each
training batch, which also speeds up convergence.
In total, we sample 14.7~million positive and 14.7~million negative
examples.
Learning is done by minimizing the $softmax$ loss \wrt~the true labels
with 500,000 iterations of standard Stochastic Gradient Descent (SGD),
with batch size 1024; starting from a base learning rate of 0.001,
that is reduced by a factor of 10 every 100,000 iterations.

\section{Experiments and Results}

To evaluate our learned similarity measure, we conduct multi-view
depth estimation experiments on several scenes. In the first
experiment, we demonstrate the performance of our approach using the
DTU~\cite{Jensen2014} evaluation framework and compare it to four
existing baseline methods.
Second, we test the similarity network's ability to generalize across
different environments. Without retraining the network, we feed it
with input images having different lighting conditions, depth range,
and sensor characteristics than what was observed during training.

The way the similarity measure is employed to provide depth map
predictions is the same for all the compared approaches. After picking
a reference view, we find the four camera
viewpoints closest to it. These five images and their camera poses are
fed to a plane-sweeping routine -- we use the publicly available code
of~\cite{Haene2014}. The explored depth range is set according to the
given dataset, and warped views are generated for $256$~uniformly
sampled disparity levels.
The SAD and ZNCC similarity measures, already implemented in the
plane-sweeping library, are used to select the best depth
estimate for each pixel position in the reference view, based on the
$4\times256$ pairwise similarity scores. To ensure a fair
comparison with the proposed approach, the same patch size of
$32\times32$ pixels is used.

For the other compared descriptors, SIFT~\cite{Lowe-IJCV-2004} and
LIFT~\cite{Simo-Serra2015,MooYi2016}, we compute similarity via the
pairwise (Euclidean) descriptor distances from the warped images, and
feed them to the same decision mechanism to select the best per-pixel
depth estimate. These descriptors were designed to work with patch
size $64\times64$ pixels, so we use these dimensions, even though it
gives them the advantage that they can base their prediction on
$4\times$ more pixels.
Note, the warping already corrects for scale and orientation, so we
skip the corresponding parts and directly compute the descriptor
entries from the warped images.
We point out that this comparison is not completely fair, since the
two descriptors were designed for sparse interest-point matching,
where some translational invariance is desirable. Still, we wanted to
test against LIFT, as the most recent learned, CNN-based descriptor.
For completeness, we include its hand-crafted counterpart.

In order to have a computationally efficient depth map prediction with
the proposed approach, we choose the input patch size to be
$128\times128$ pixels. This allows the network to compute the
similarity scores for $25\times25$ partially overlapping patches
(stride $4$) of size $32\times32$ in a single forward pass, filling up
a region of $100\times100$ similarity scores (after upsampling). Doing
this for every depth of the sweeping plane, we obtain a list of $256$
multi-view similarity scores per pixel, and simply pick the depth with
the highest score.  To compute the $25\times25\times256$ similarity
scores (covering an area of $100\times100$~original pixels) takes
$70\,\mathrm{ms}$, on an Intel~i7 computer with Nvidia Titan~X GPU
using Caffe in Linux.

Practical matching software does not return raw depth estimates, but
improves them with different simple steps. The plane-sweeping
library~\cite{Haene2014} offers two such options: (i) interpolation of
the discrete depth levels to sub-pixel accuracy, and (ii) box
filtering to account for correlations between nearby depth values. We
tested both options and found them to consistently improve the overall
results, independent of the similarity measure. As we are primarily
interested how well different similarity metrics perform under
realistic conditions, we enable subpixel accuracy and box filtering in
all following experiments.

\subsection{Evaluation on the DTU dataset}

The DTU Robot Image Data (DTU) contains more than 80~different indoor
table top scenes. Each DTU scene was recorded from predetermined,
known camera poses with the help of a camera mounted on a robot arm,
and a ground truth 3D~point cloud was generated by mounting a
high-accuracy active stereo sensor on the same robot arm. Images from
49~scenes are already used as our training data. Beyond those, also
scenes that share one or more objects with any of the training scenes
are excluded from testing, to avoid potential biases. For DTU, we set
the depth range to $[0.45\hdots1]\,\mathrm{m}$.

\vspace{-0.5em}
\paragraph{Qualitative Results.}

The four example objects we use in the following are labeled BIRD,
FLOWER, CAN, and BUDDHA. The BIRD has a uniform color so the intensity
differences are largely due to the shading effects. The FLOWER is a
difficult object due to the thin structure and the small
leaves. Underneath the flowerpot, there is a newspaper with strong
texture. The CAN has a metallic surface with homogeneous texture,
while the background is rather strongly textured. The BUDDHA is made
out of porcelain, with significant specular reflections.

The provided color images are converted to grayscale before
processing, \cf Fig.~\ref{fig:img1}-\ref{fig:img4}. The ground truth
depth maps have been generated by back-projecting ground truth 3D~point
clouds and are sparse, \cf Fig.~\ref{fig:gt1}-\ref{fig:gt4}, whereas
the depth maps from multi-view stereo are dense.  We thus evaluate
only at the sparser ground truth depths. Also the difference plots in
Fig.~\ref{fig:our1}-\ref{fig:lift4} show only the pixels for which
ground truth is available.
Depth differences are color-coded on a logarithmic scale, where dark
blue corresponds to zero difference and dark red denotes differences
$>20\,\mathrm{mm}$.

\makeatletter
\definecolor{darkblue}{rgb}{0,0,0.8}
\definecolor{darkred}{rgb}{0.55,0,0}
\makeatother
\begin{figure*}[thp]
\centering \setlength{\tabcolsep}{0pt}
\begin{tabular}{cccc}
   \subfloat[]{
     \includegraphics[width=0.5\columnwidth]{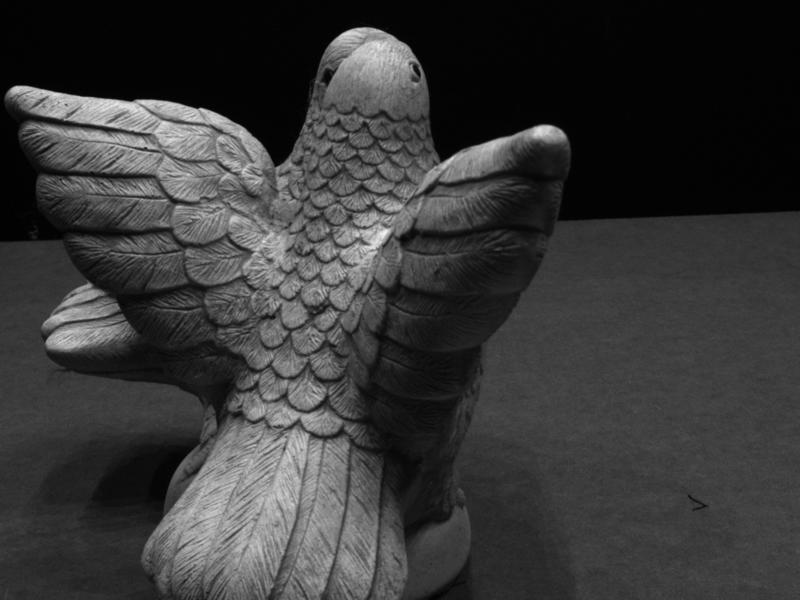}
     \label{fig:img1}} &
   \subfloat[]{
     \includegraphics[width=0.5\columnwidth]{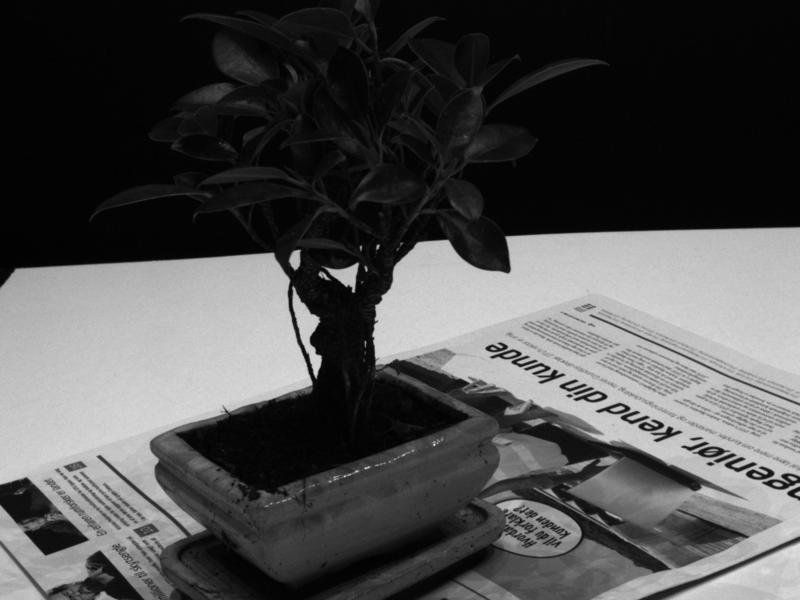}
     \label{fig:img2}} &
   \subfloat[]{
     \includegraphics[width=0.5\columnwidth]{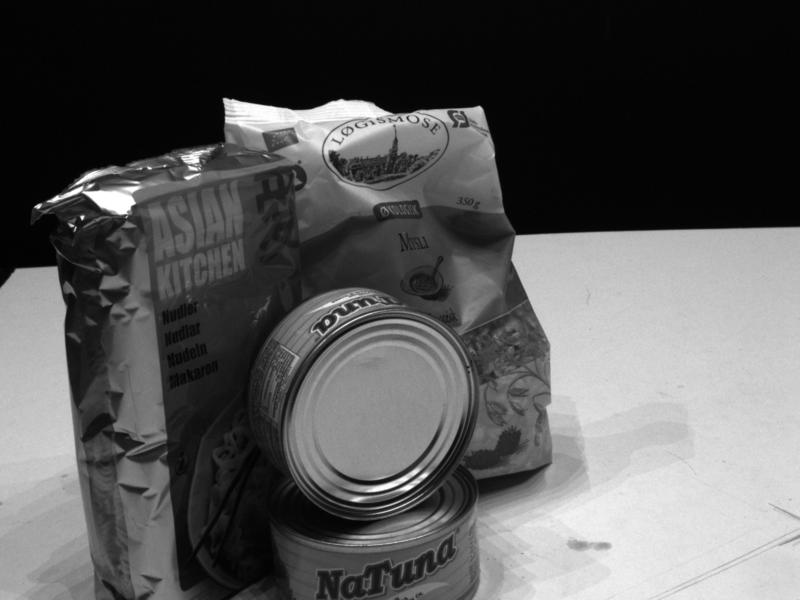}
     \label{fig:img3}} &
   \subfloat[]{
     \includegraphics[width=0.5\columnwidth]{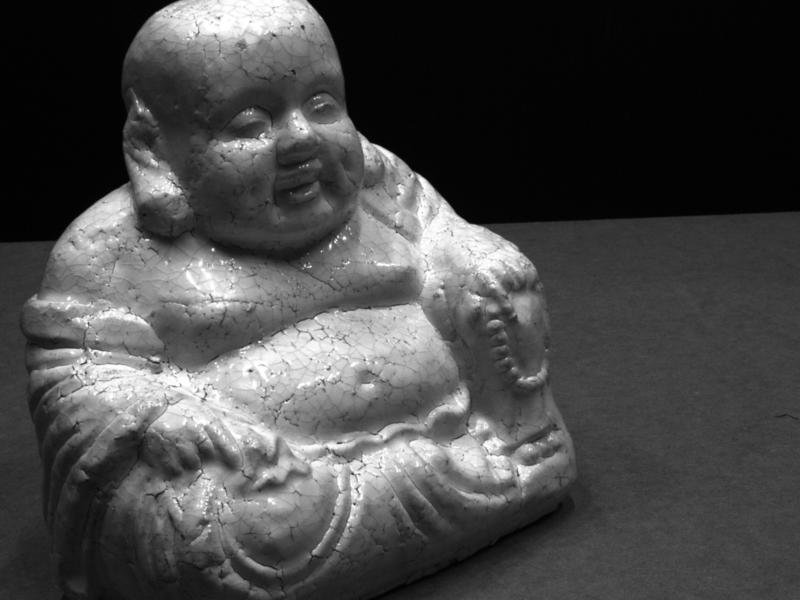}
   \label{fig:img4}} \\
   \vspace{-0.7em}
   \subfloat[]{
     \includegraphics[width=0.5\columnwidth]{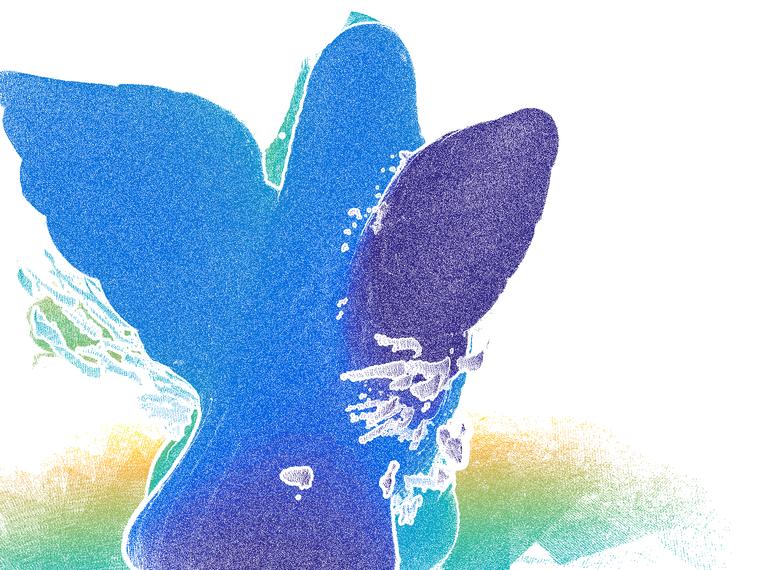}
     \label{fig:gt1}} &
   \subfloat[]{
     \includegraphics[width=0.5\columnwidth]{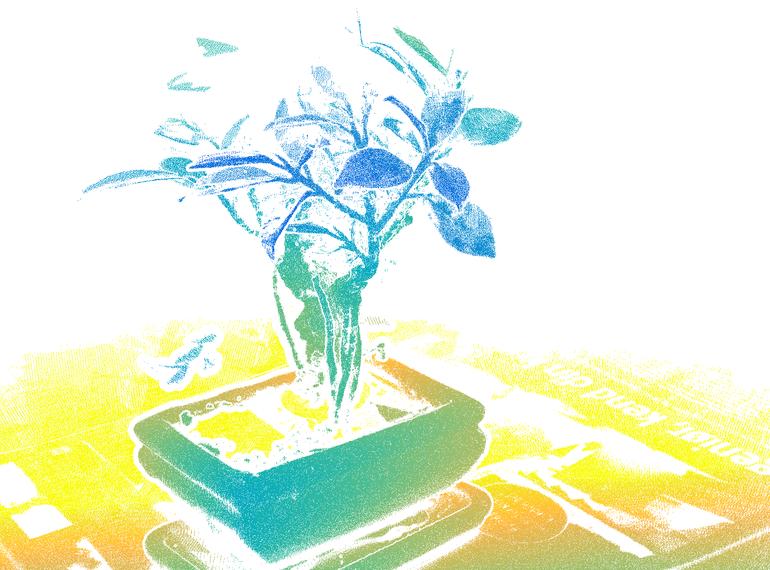}
     \label{fig:gt2}} &
   \subfloat[]{
     \includegraphics[width=0.5\columnwidth]{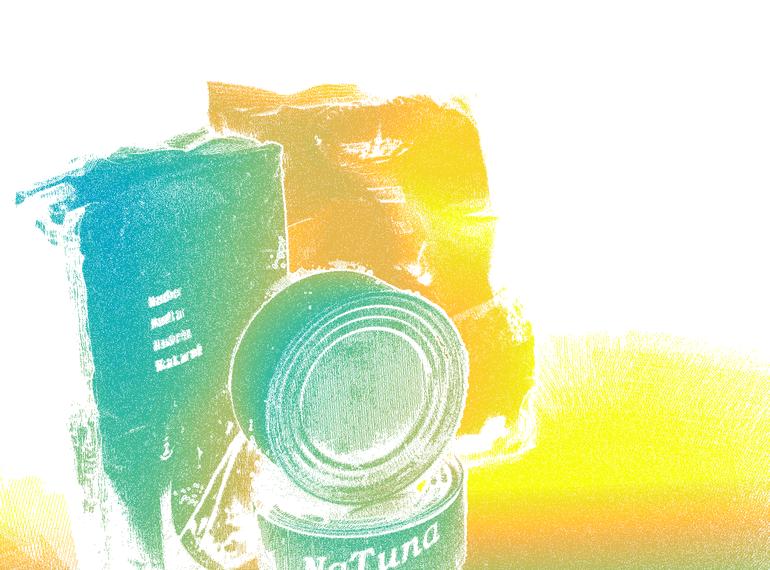}
     \label{fig:gt3}} &
   \subfloat[]{
     \includegraphics[width=0.5\columnwidth]{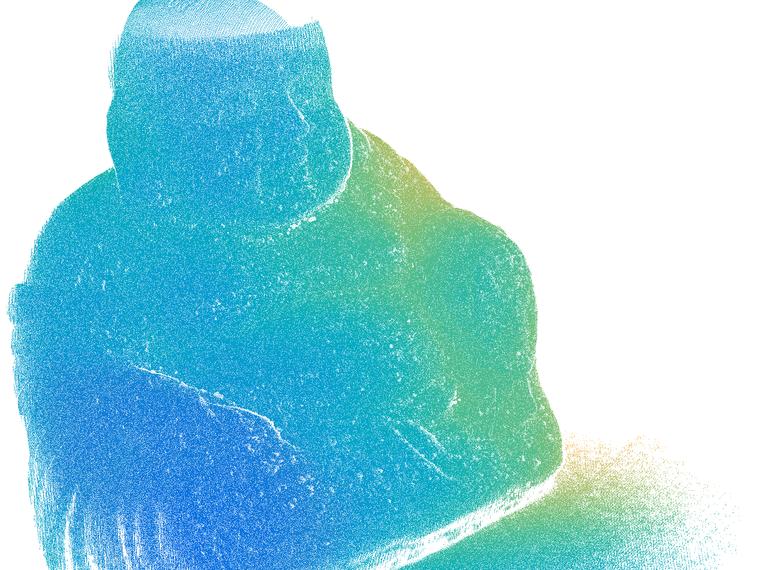}
     \label{fig:gt4}} \\
   \vspace{-0.7em}
   \subfloat[]{
     \includegraphics[width=0.5\columnwidth]{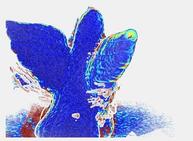}
     \label{fig:our1}} &
   \subfloat[]{
     \includegraphics[width=0.5\columnwidth]{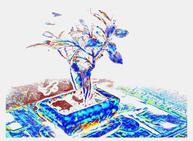}
     \label{fig:our2}} &
   \subfloat[]{
     \includegraphics[width=0.5\columnwidth]{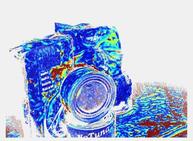}
     \label{fig:our3}} &
   \subfloat[]{
     \includegraphics[width=0.5\columnwidth]{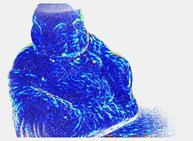}
     \label{fig:our4}} \\
   \vspace{-0.7em}
   \subfloat[]{
     \includegraphics[width=0.5\columnwidth]{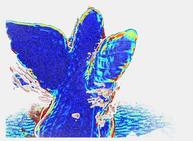}
     \label{fig:zncc1}} &
   \subfloat[]{
     \includegraphics[width=0.5\columnwidth]{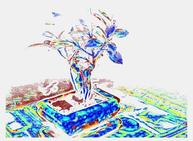}
     \label{fig:zncc2}} &
   \subfloat[]{
     \includegraphics[width=0.5\columnwidth]{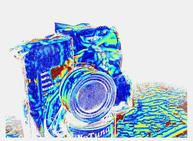}
     \label{fig:zncc3}} &
   \subfloat[]{
     \includegraphics[width=0.5\columnwidth]{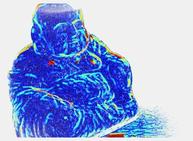}
     \label{fig:zncc4}} \\
   \vspace{-0.7em}
   \subfloat[]{
     \includegraphics[width=0.5\columnwidth]{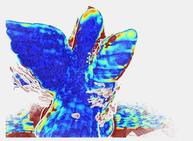}
     \label{fig:lift1}} &
   \subfloat[]{
     \includegraphics[width=0.5\columnwidth]{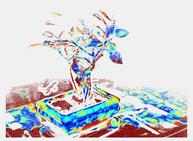}
     \label{fig:lift2}} &
   \subfloat[]{
     \includegraphics[width=0.5\columnwidth]{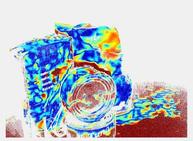}
     \label{fig:lift3}} &
   \subfloat[]{
     \includegraphics[width=0.5\columnwidth]{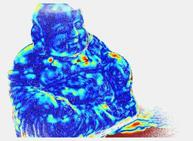}
     \label{fig:lift4}}
\end{tabular}
\vspace{-0.3em}
\caption{Plane sweeping using three different patch similarity
  measures. Proposed learned multi-view similarity \vs pairwise ZNCC
  and pairwise LIFT. Reference images (a-d). Ground truth
  (e-h). Absolute deviations from ground truth for the proposed method
  (i-k), ZNCC (m-p) and LIFT (q-t). \textcolor{blue}{Blue}: no
  difference, \textcolor{darkred}{red}: high difference.
          }
\label{fig:results}
\end{figure*}

For the BIRD the right wing, which is partly in shadow, is the most
difficult part. For all three methods there are errors at the wing
boundary, the largest errors are observed for LIFT. Note also errors
on the left wing boundary present in ZNCC and LIFT but not in the
proposed method.
The most difficult object is the FLOWER. Here the ZNCC and our
approach again outperform LIFT. All three methods
struggle with the dark, homogeneous soil in the flowerpot. On the
leaves, as far as present in the ground truth, our method has the
smallest error, especially near the boundaries.
For the CAN object, the homogeneous metal surface challenges the
LIFT approach, whereas the two others can resolve it
correctly.
For the BUDDHA the most difficult parts for depth estimation are the
small regions with specular reflection. These can be seen
in Figs.~\ref{fig:teaser},~\ref{fig:img4}. In those regions
ZNCC and LIFT have large errors, while our direct multi-view approach
copes a lot better (Fig.~\ref{fig:our4}).
Overall, the examples suggest that our proposed similarity is
more general: the classic ZNCC method often works well, but
struggles near object boundaries. LIFT seems to have problems with
certain surfaces materials like the metal of the CAN, and both
competitors struggle with specularities (which, incidentally, are a
main problem of stereo reconstruction, especially in industrial
applications).

\vspace{-0.5em}
\paragraph{Quantitative results.}

The authors of the DTU dataset provide an evaluation framework in
which, per object, the groundtruth 3D~point cloud is compared to the
one constructed from the depth estimates.
For the evaluation we thus convert the
estimated depth maps to 3D~point clouds by lifting them with the known
camera calibration. Note that we do not use
depthmap integration across multiple viewpoints.
\emph{Accuracy} is defined as the average (truncated) distance from a
reconstructed point to the nearest ground truth
point. \emph{Completeness} is defined inversely, as average distance
from ground truth points to the nearest model points, \ie, lower
values are also better.

\begin{table}[t]
  \small
  \begin{center}
  \begin{tabular}{|l|c|cc|cc|}\hline
  Similarity & \multicolumn{1}{c|}{\#views} & \multicolumn{2}{c|}{~Accuracy~} & \multicolumn{2}{c|}{~Completeness~}\\
  ~~~~~~~~~~~~~~~~~~~ & & Mean & Median & Mean & Median\\\hline
  SAD & 4$\times$2 & 1.868 & 0.617 & 2.534 & 1.148 \\
  ZNCC & 4$\times$2 & {\bf 1.237} & 0.577 & 2.260 & 1.025 \\
  {\itshape OUR stereo} & 4$\times$2 & 1.339 & {\bf 0.453} & 2.964 & 1.391 \\ 
  {\itshape OUR 5-view} & 5 & 1.356 & 0.472 & {\bf 2.126} & {\bf 0.868} \\ \hline
  \end{tabular}
  \end{center}
  \vspace{-.5em}
  \caption{Quantitative results for 20 objects from the DTU
    datasets. Four similarity measures are compared: sum of absolute
    differences, zero-mean normalized cross correlation, proposed
    similarity over 2 views, and proposed similarity over 5 views.}
  \label{tab:quantit_stereo}
\end{table}

We start with a comparison to standard similarity measures for 20
scenes in Table~\ref{tab:quantit_stereo}. While the mean values
describe the overall performance, median values are less sensitive to
outliers. Note that, also for the mean, the per-point distances of
outliers are truncated at $20\,\mathrm{mm}$.
To support the claim that direct multi-view similarity is preferable
to consensus over 2-view scores, we also run our method in 2-view
stereo mode and feed the scores to the same consensus mechanism used
for SAD and ZNCC. Directly computing similarity over 5 input views
delivers significantly better completeness at similar accuracy as the
best competitors.
We did not run the full experiment with LIFT, due to the excessive
runtime of pixelwise CNN prediction without a fully convolutional
architecture.

\begin{table}[t]
  \small
  \begin{center}
  \begin{tabular}{|l|cc|cc|}\hline
  Similarity & \multicolumn{2}{c|}{~Accuracy~} & \multicolumn{2}{c|}{~Completeness~}\\
  (5 views) & Mean & Median & Mean & Median\\\hline
  \multicolumn{1}{|c}{} & \multicolumn{4}{c|}{~BIRD~} \\\hline
  SAD & 2.452 & 0.380 & {\bf 4.035} & 1.105\\
  ZNCC & {\bf 1.375} & 0.365 & 4.253 & 1.332\\
  SIFT & 1.594 & 0.415 & 5.269 & 1.845\\
  LIFT & 1.844 & 0.562 & 4.387 & 1.410\\
  {\itshape OUR concat} & 1.605 & 0.305 & 4.358 & 1.133\\
  {\itshape OUR} & 1.881 & {\bf 0.271} & 4.167 & {\bf 1.044}\\\hline
  \multicolumn{1}{|c}{} & \multicolumn{4}{c|}{~FLOWER~} \\\hline
  SAD & 2.537 & 1.143 & 2.768 & 1.407\\
  ZNCC & 2.018 & 1.106 & 2.920 & 1.467\\
  SIFT & 2.795 & 1.183 & 4.747 & 2.480\\
  LIFT & 3.049 & 1.420 & 4.224 & 2.358 \\
  {\itshape OUR concat} & 2.033 & 0.843 & \bf{2.609} & 1.267\\
  {\itshape OUR} & {\bf 1.973} & {\bf 0.771} & {\bf 2.609} & {\bf 1.208}\\\hline
  \multicolumn{1}{|c}{} & \multicolumn{4}{c|}{~CAN~} \\\hline
  SAD & 1.824 & 0.664 & 2.283 & 1.156\\
  ZNCC & 1.187 & 0.628 & 2.092 & 1.098\\
  SIFT & 1.769 & 0.874 & 3.067 & 1.726\\
  LIFT & 2.411 & 1.207 & 3.003 & 1.823\\
  {\itshape OUR concat} & {\bf 1.082} & {\bf 0.477} & {\bf 1.896} & {\bf 0.833}\\
  {\itshape OUR} & 1.123 & 0.478 & 1.982 & 0.874\\\hline
  \multicolumn{1}{|c}{} & \multicolumn{4}{c|}{~BUDDHA~} \\\hline
  SAD & 0.849 & 0.250 & 1.119 & 0.561\\
  ZNCC & 0.688 & 0.299 & 1.208 & 0.656\\
  SIFT & 0.696 & 0.263 & 1.347 & 0.618 \\
  LIFT & 0.688 & 0.299 & 1.208 & 0.656\\
  {\itshape OUR concat} & 0.682 & 0.231 & {\bf 1.017} & {\bf 0.473}\\
  {\itshape OUR} & {\bf 0.637} & {\bf 0.206} & 1.057 & 0.475\\\hline
  \end{tabular}
  \end{center}
  \vspace{-0.5em}
  \caption{Quantitative results for BIRD, FLOWER, CAN, and BUDDHA objects. Six
    similarity measures are compared: sum of absolute differences, zero mean normal cross correlation, SIFT, LIFT, proposed
    multi-view similarity using concatenation,
    and proposed multi-view similarity using averaging.}
  \label{tab:quantit}
  \end{table}

In Table~\ref{tab:quantit} we compare the accuracy and completeness
of all tested methods for the four example scenes.
Differences are relatively small between SAD and ZNCC, probably due to
the controlled lighting.
The results for SIFT and LIFT are also quite similar, except for the
CAN object where SIFT clearly outperforms its learned counterpart.

The proposed method achieves the best median accuracy and median
completeness in all the scenes, and the best mean accuracy and
completeness for three of them. Note that there is virtually no
difference between averaging and concatenation. There seems to be no
performance penalty for averaging, while at the same time one gains
the flexibility to use a variable number of input views.
On the BIRD, our method ranks third in accuracy and second in
completeness. There are relatively big differences between median and
mean errors, apparently all measures show quite good performance on
the rather diffuse surface, whereas outliers due to ambiguous texture
inflate the mean values.

Overall, the proposed multi-view patch similarity exhibits the
quantitatively and qualitatively best performance. In particular, the
experiments support our claim that learning end-to-end multi-view
similarity is preferable to comparing learned per-patch descriptors
with conventional Euclidean distance, and to a consensus over learned
2-view similarities.

\makeatletter
\newlength{\halfcol}
\setlength\halfcol{0.48\columnwidth}
\makeatother
\begin{figure}[t]
\centering \setlength{\tabcolsep}{0pt}
\begin{tabular}{cc}
 \subfloat[input view]{
   \includegraphics[width=\halfcol]{figures/results/reference_image107.jpg}
   \label{fig:orig}} &
 \subfloat[3 views]{
   \includegraphics[width=\halfcol]{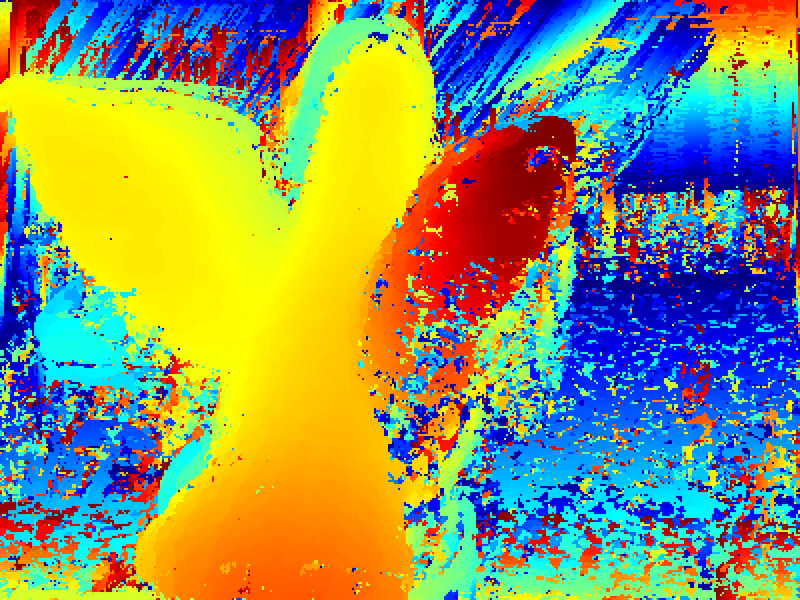}
 \label{fig:threev}} \\[-0.3ex]
    \subfloat[5 views]{
   \includegraphics[width=\halfcol]{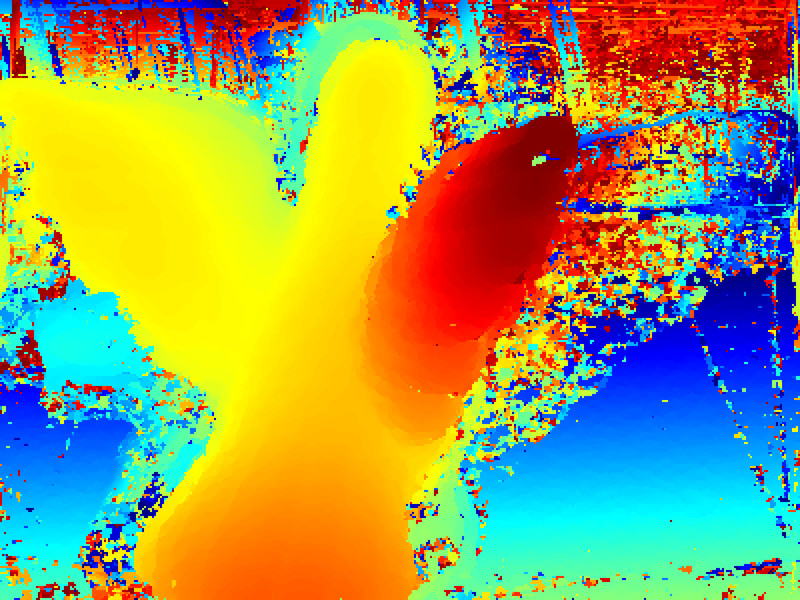}
   \label{fig:fivev}} &
 \subfloat[9 views]{
   \includegraphics[width=\halfcol]{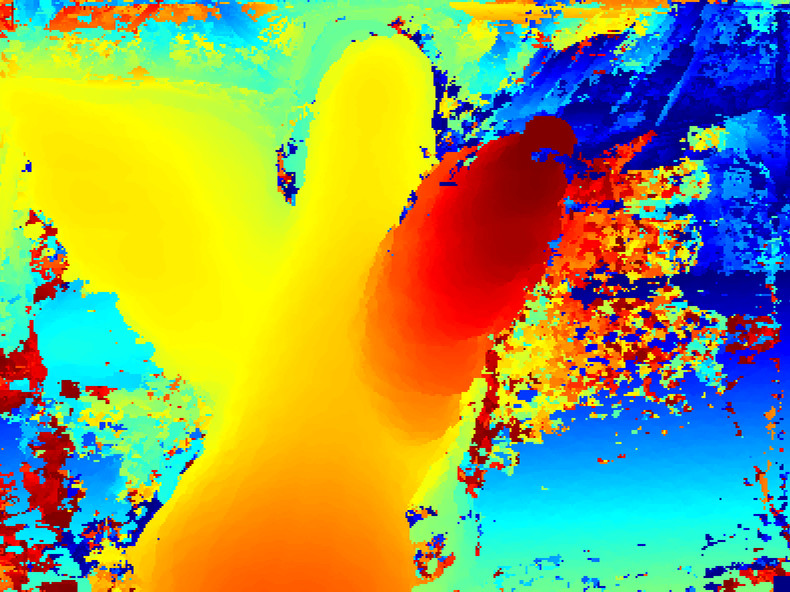}
   \label{fig:ninev}}
\end{tabular}
\caption{Matching different numbers of views with the similarity
  network can be done without retraining. Result displayed without
  sub-pixel refinement and box filtering to accentuate differences.}
\label{fig:diffnov}
\end{figure}

\paragraph{Varying the number of input patches.}
We go on to demonstrate that the network architecture, in which
branches are averaged, can be applied to different numbers of input
views without retraining. We run experiments with either three or
nine views, respectively patches. Both give reasonable depth
estimates (Fig.~\ref{fig:diffnov}). As expected, the {results} with only three views are a bit
worse. However, using nine patches instead of five further improves
performance -- although the similarity network has only been trained
with five. We speculate that information how to correctly weight
pixels at discontinuities, occlusions etc.\ passes down the individual
branches during training, so that the parallel Siamese streams can
also use it for additional branches. On the other hand, averaging
itself may have a stronger denoising effect with more
branches. Further research is needed to clarify the underlying
mechanisms.
Whatever the reason, in our view flexibility with respect to the
number of input views is an attractive and important feature for
real-world applications of multi-view stereo.

\subsection{Evaluation on the Fountain dataset}

The images recorded with the robot for the DTU dataset are all taken
in an indoor laboratory environment, and one would expect that the
specific lighting and camera characteristics of the dataset are
captured in our trained network. Therefore, we apply the learned
multi-view similarity measure also to the well-known Fountain
dataset~\cite{strecha2008}, without retraining it.

\begin{figure}[t]
\centering \setlength{\tabcolsep}{0pt}
\begin{tabular}{cc}
 \subfloat[ground truth]{
   \includegraphics[width=\halfcol]{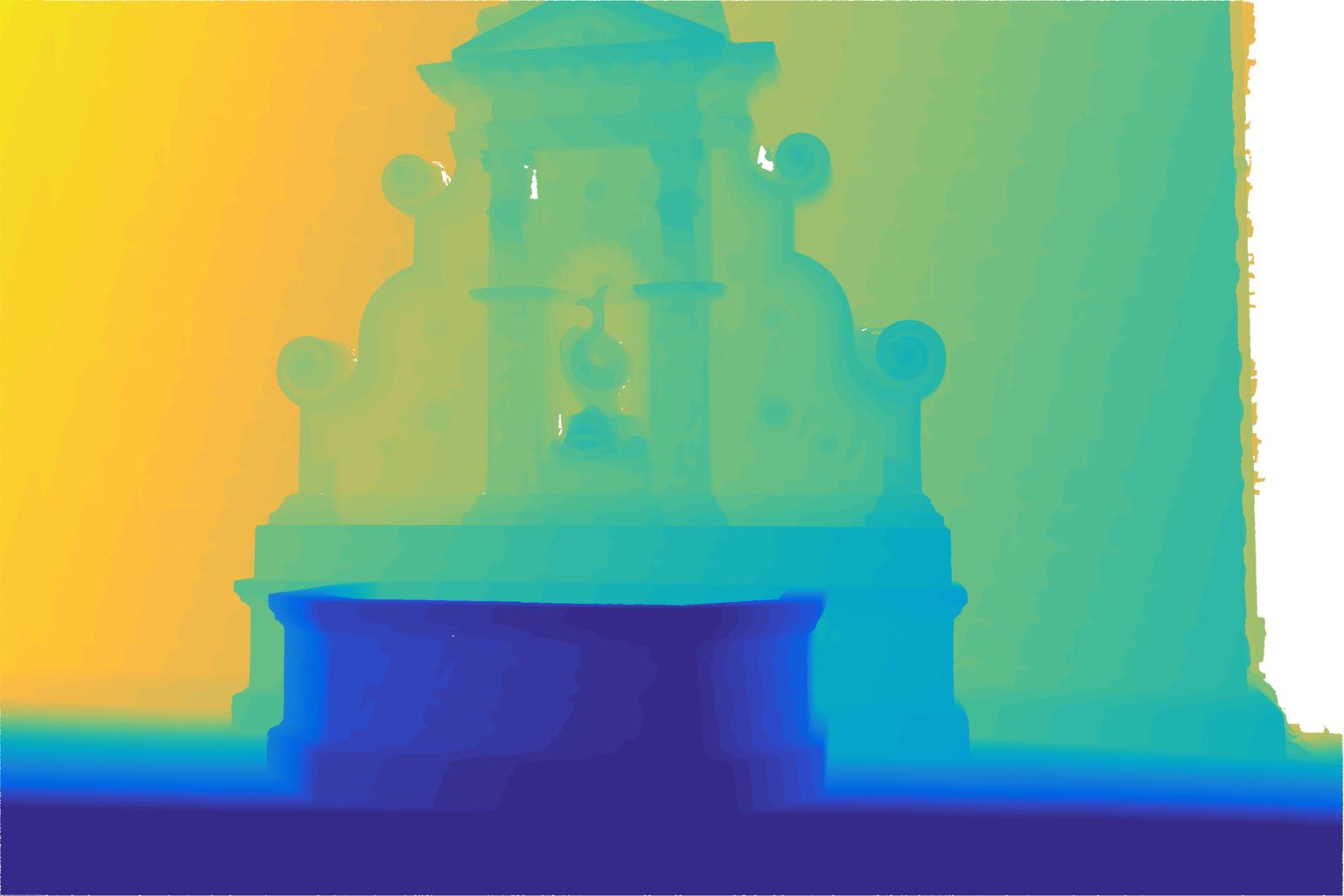}
   \label{fig:imGTf}} &
 \subfloat[OUR]{
   \includegraphics[width=\halfcol]{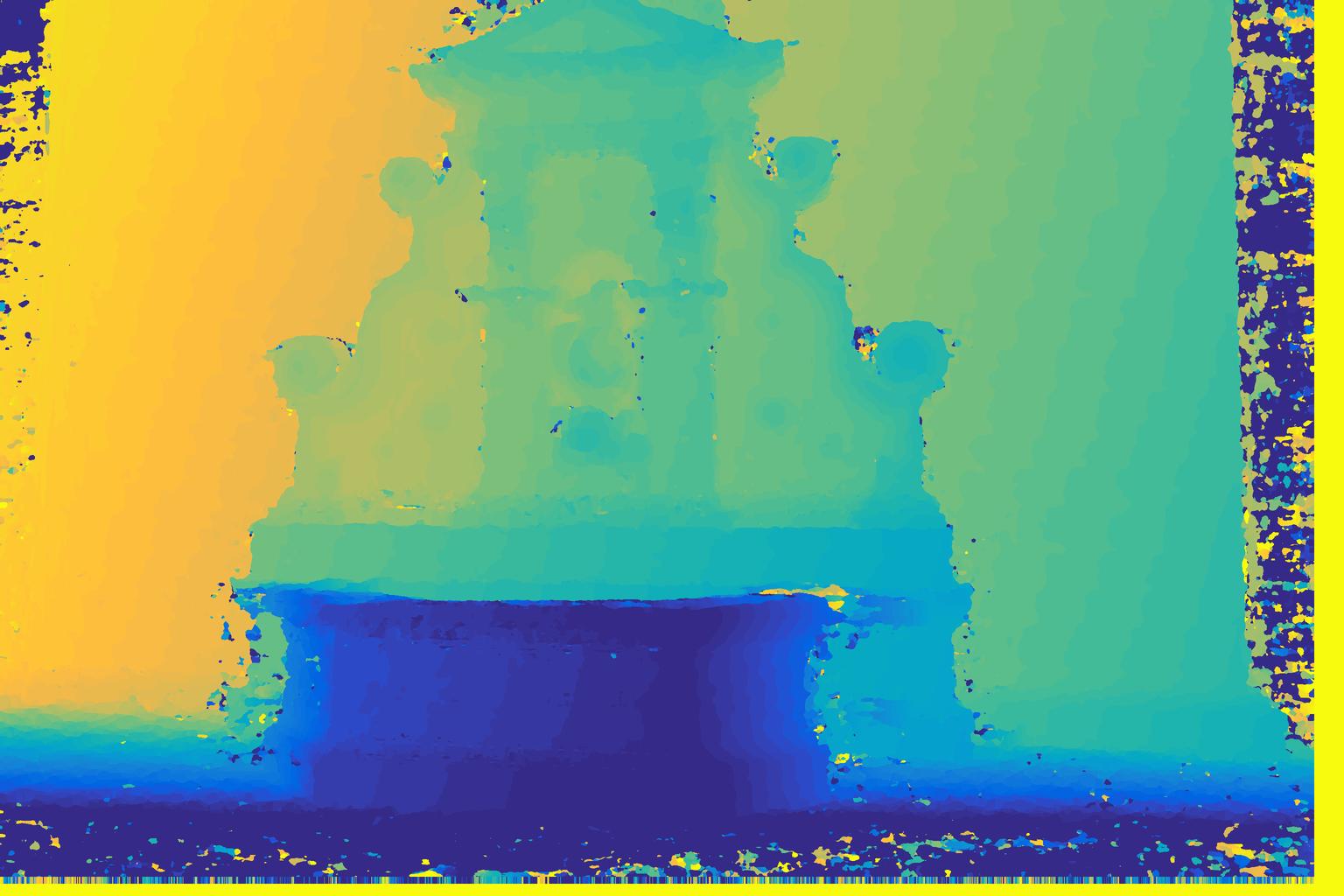}
   \label{fig:fount_OUR}} \\[-0.7ex]
    \subfloat[SIFT]{
   \includegraphics[width=\halfcol]{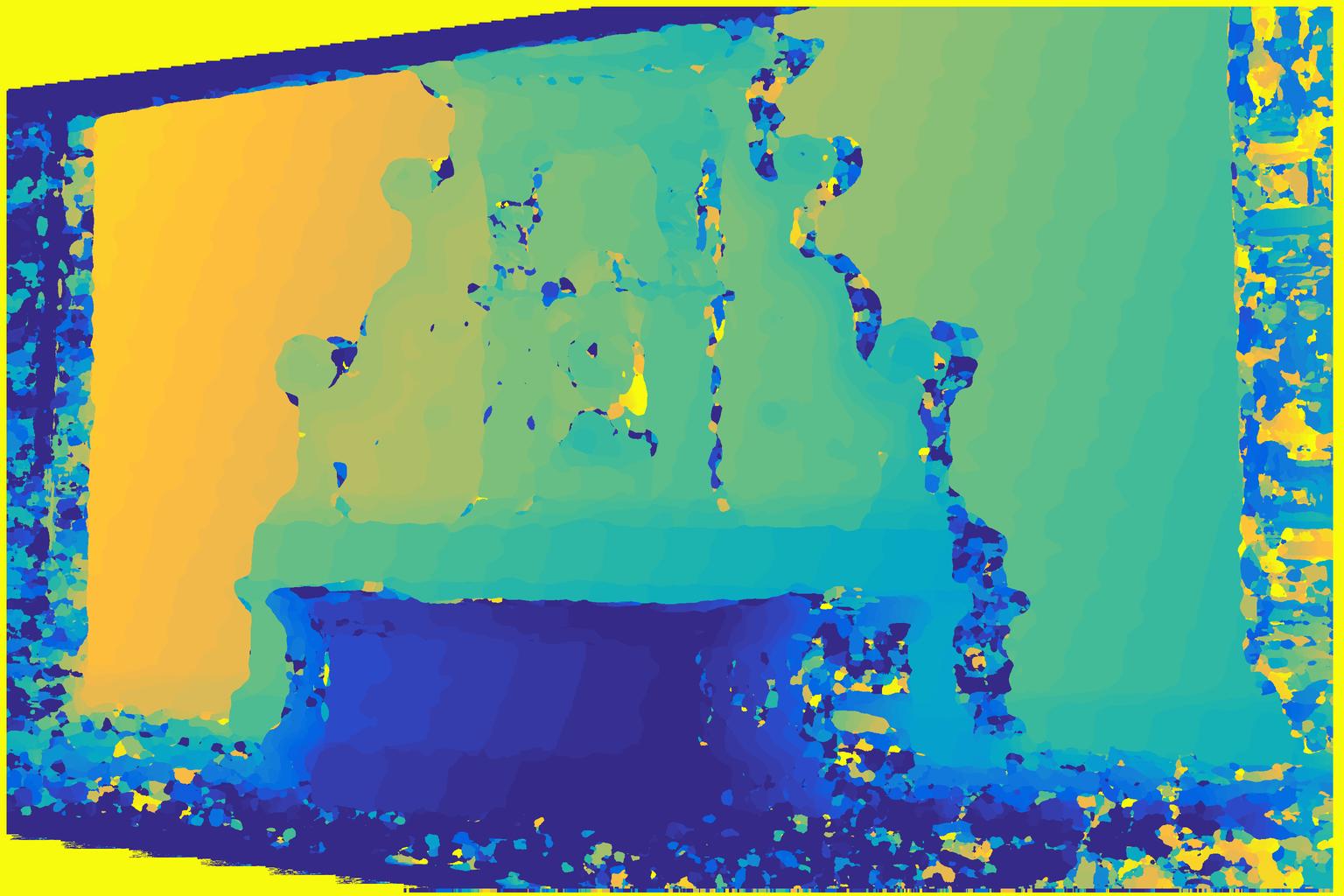}
   \label{fig:fount_SIFT}} &
 \subfloat[ZNCC]{
   \includegraphics[width=\halfcol]{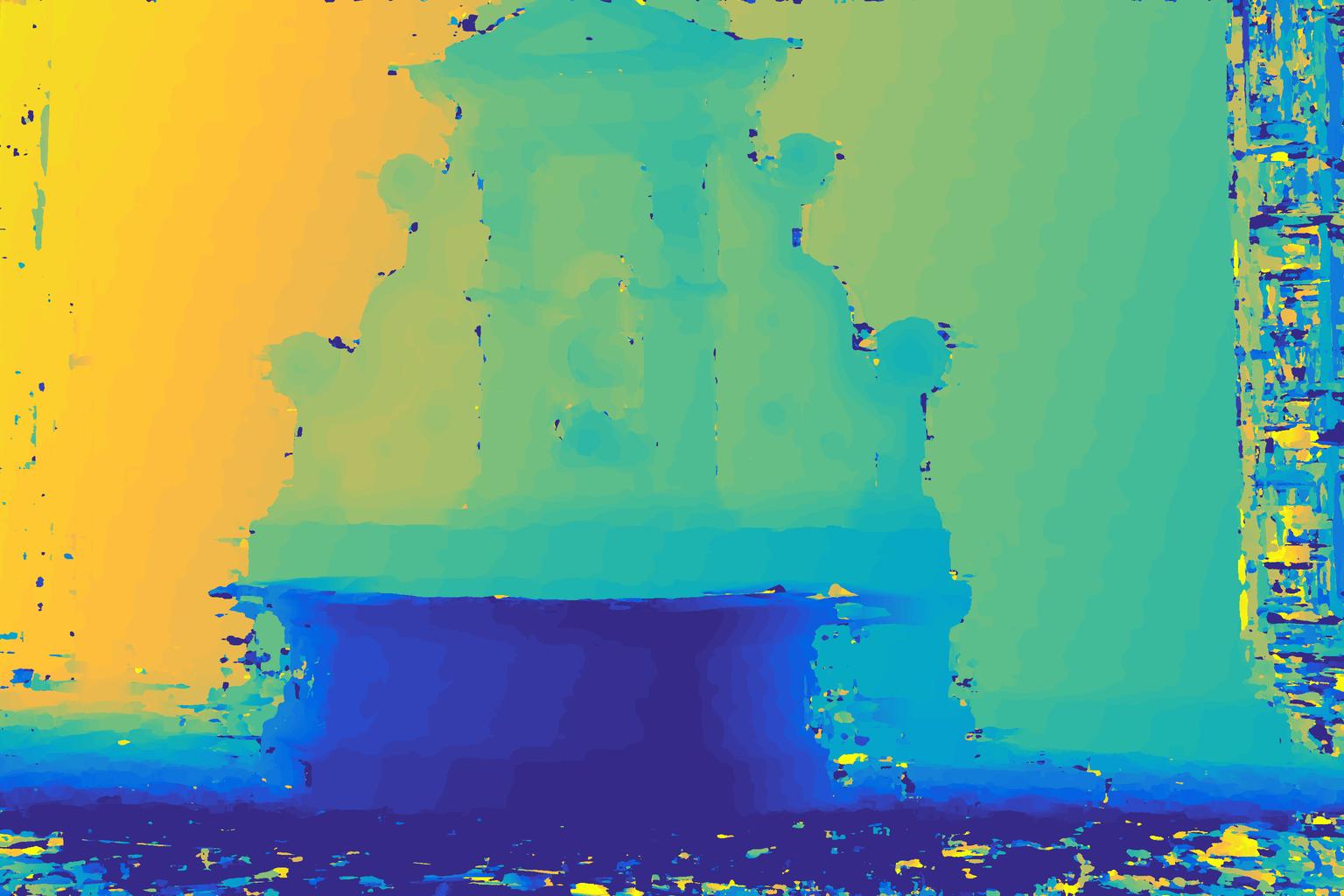}
   \label{fig:fount_ZNCC}} \\
\end{tabular}
\caption{The learned similarity generalizes to a different test
  environment, seemingly as well as the competing descriptors.}
\label{fig:fountain}
\vspace{-0.8em}
\end{figure}

For the experiment we select five neighboring images and set the depth
range to $[5\hdots10]\,\mathrm{m}$. The depth maps in
Fig.~\ref{fig:fountain} show the qualitative result for three
different methods. Our method works at least as well as ZNCC and SIFT,
which are generic and not learned from data. The network does not seem
to significantly overfit to the DTU setting.

\section{Conclusion}

We have proposed to directly learn a similarity / matching score over
a set of \emph{multiple} patches with a discriminative learning
engine, rather than heuristically assemble it from pairwise
comparisons.
An $n$-way Siamese convolutional network, which applies the same,
learned transformation to all input patches and combines the results,
was identified as a suitable state-of-the-art learning engine.

From a high-level perspective, machine learning for 3D~vision started
with very small steps like learning the distance metric between given
descriptors, or learning descriptor extraction for pairwise matching.
Recently, bolder ideas have emerged, all the way to learning an
end-to-end mapping from images to (volumetric, low-resolution)
3D~models~\cite{choy16}.
Our work is positioned somewhere in the middle. While we do not see an
immediate reason to replace the geometrically transparent and
well-understood structure-from-motion pipeline with a learned
black-box predictor, we do agree that certain steps of the pipeline
lack a principled solution and might be best addressed with machine
learning.
In our work, we have used the learned similarity score for multi-view
stereo matching, but we believe that variants of it could also be
beneficial for other applications where similarity, respectively
distance, between more than two images must be assessed.

\vspace{-0.7em}
\paragraph*{Dedication.}
We dedicate this paper to the memory of Wilfried Hartmann, who did not
live to see it published.

\vspace{-0.8em}
\paragraph*{Acknowledgements.}
The work was funded in part by the EU's Horizon 2020 programme under
grant agreement 687757 -- REPLICATE, and supported by NVIDIA
Corporation through an Academic Hardware Grant. We thank D.~Marmanis
and K.K.~Maninis for valuable comments.

{\small
\bibliographystyle{ieee}
\bibliography{mylit,mybib}
}

\newpage

\begin{table*}
\section*{Supplementary material}
\subsection*{Qualitative Results - DTU Evaluation}
The qualitative results on the DTU dataset are presented in
(Fig.~\ref{fig:results}) in the
submitted paper. There the difference to ground truth depth maps are shown for
three out of six methods. In (Fig.~\ref{fig:results_diff2}) the results for the
remaining three methods are shown.

The methods SAD and SIFT perform different on the CAN dataset compared to the
ZNCC and LIFT results. SAD has difficulties with the homogeneous can surface.
For SIFT there are more correct depth estimates on the homogeneous can surface
when comparing to LIFT. For the other three datasets there are no large
differences seen when comparing the two proposed methods, ZNCC with SAD and LIFT
with SIFT.
\end{table*}
\begin{figure*}[t]
\centering \setlength{\tabcolsep}{0pt}
\begin{tabular}{cccc}
   \subfloat[]{
     \includegraphics[width=0.24\textwidth]{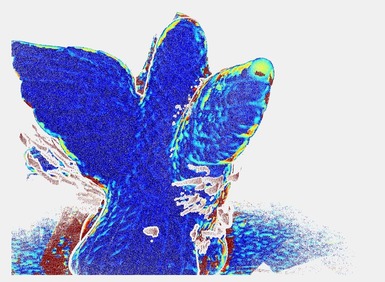}
     \label{fig:our1}} &
   \subfloat[]{
     \includegraphics[width=0.24\textwidth]{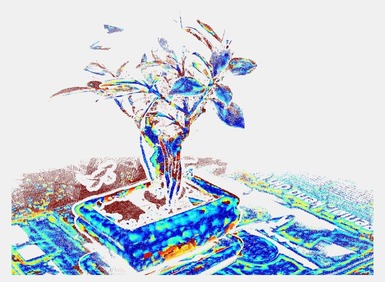}
     \label{fig:our2}} &
   \subfloat[]{
     \includegraphics[width=0.24\textwidth]{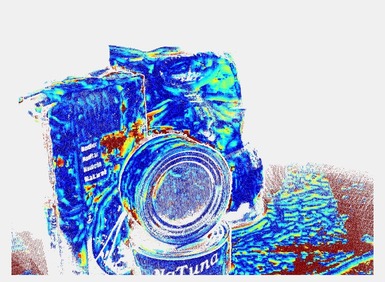}
     \label{fig:our3}} &
   \subfloat[]{
     \includegraphics[width=0.24\textwidth]{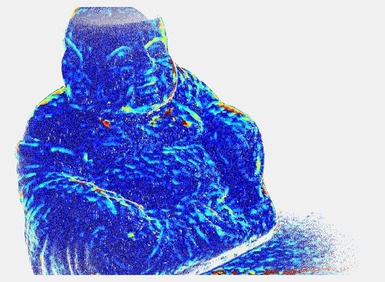}
     \label{fig:our4}} \\
   \subfloat[]{
     \includegraphics[width=0.24\textwidth]{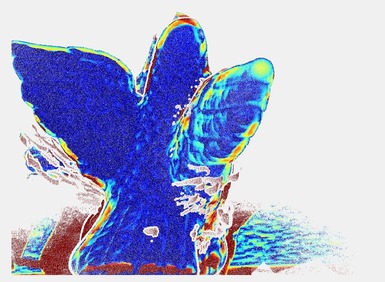}
     \label{fig:zncc1}} &
   \subfloat[]{
     \includegraphics[width=0.24\textwidth]{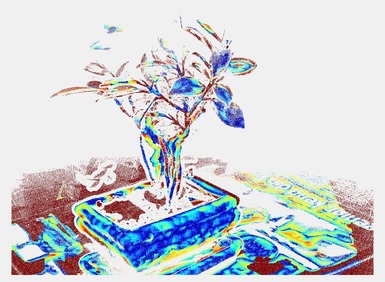}
     \label{fig:zncc2}} &
   \subfloat[]{
     \includegraphics[width=0.24\textwidth]{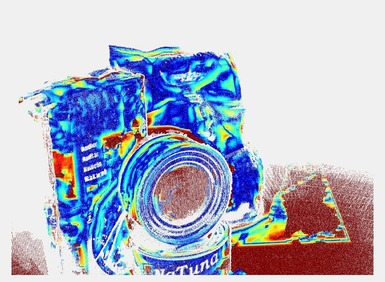}
     \label{fig:zncc3}} &
   \subfloat[]{
     \includegraphics[width=0.24\textwidth]{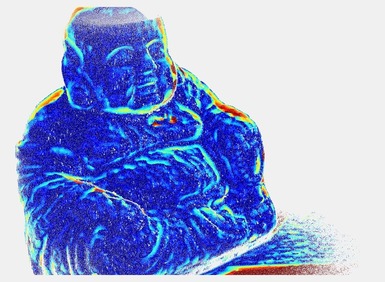}
     \label{fig:zncc4}} \\
   \subfloat[]{
     \includegraphics[width=0.24\textwidth]{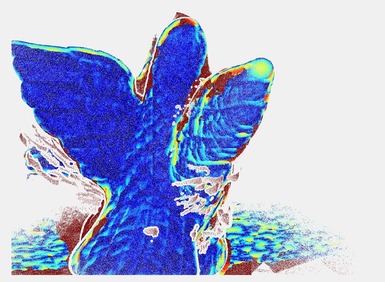}
     \label{fig:lift1}} &
   \subfloat[]{
     \includegraphics[width=0.24\textwidth]{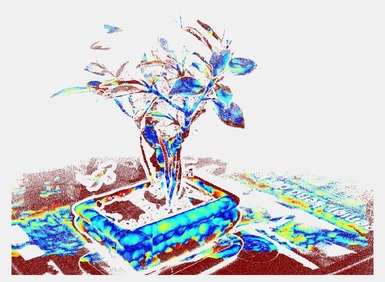}
     \label{fig:lift2}} &
   \subfloat[]{
     \includegraphics[width=0.24\textwidth]{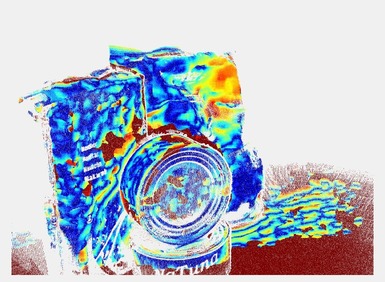}
     \label{fig:lift3}} &
   \subfloat[]{
     \includegraphics[width=0.24\textwidth]{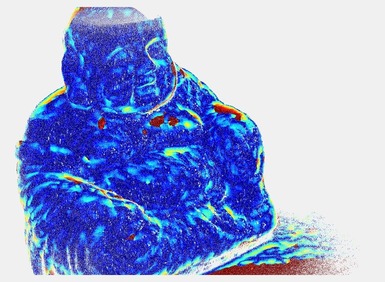}
     \label{fig:lift4}} \\
\end{tabular}
\caption{Plane sweeping using three different patch similarity
  measures. Proposed learned multi-view similarity with concatenation layer \vs pairwise SAD
  and pairwise SIFT. (a) - (d) difference from ground truth, proposed. (e) - (h)
  difference from ground truth, SAD. (i) - (l) difference from ground
  truth, SIFT.}
\label{fig:results_diff2}
\end{figure*}
\end{document}